\def\year{2022}\relax
\documentclass[letterpaper]{article} % DO NOT CHANGE THIS
\usepackage{aaai22}  % DO NOT CHANGE THIS
\usepackage{times}  % DO NOT CHANGE THIS
\usepackage{helvet}  % DO NOT CHANGE THIS
\usepackage{courier}  % DO NOT CHANGE THIS
\usepackage[hyphens]{url}  % DO NOT CHANGE THIS
\usepackage{graphicx} % DO NOT CHANGE THIS
\usepackage{natbib}  % DO NOT CHANGE THIS AND DO NOT ADD ANY OPTIONS TO IT
\usepackage{caption} % DO NOT CHANGE THIS AND DO NOT ADD ANY OPTIONS TO IT
\DeclareCaptionStyle{ruled}{labelfont=normalfont,labelsep=colon,strut=off} % DO NOT CHANGE THIS
\usepackage{algorithm}
\usepackage{algorithmic}

\newcommand{\ie}{\textit{i}.\textit{e}.}
\newcommand{\eg}{\textit{e}.\textit{g}.}
\usepackage{amsmath}
\usepackage{amssymb}
\usepackage{makecell}
\usepackage{multirow}
\usepackage{multicol}
\usepackage{booktabs}
\usepackage{pifont}
\newcommand{\cmark}{\ding{51}}
\newcommand{\xmark}{\ding{55}}
\usepackage{arydshln}
\usepackage[dvipsnames]{xcolor}

\usepackage{newfloat}
\usepackage{listings}
\floatstyle{ruled}
\newfloat{listing}{tb}{lst}{}
\floatname{listing}{Listing}
\title{Distinguishing Homophenes Using Multi-Head \\ Visual-Audio Memory for Lip Reading}
\author{
    Minsu Kim, Jeong Hun Yeo, Yong Man Ro\thanks{Corresponding author.}
}
\affiliations{
    Image and Video Systems Lab, KAIST, South Korea\\
    \{ms.k, sedne246, ymro\}@kaist.ac.kr
}

\begin{document}

\maketitle

\begin{abstract}
Recognizing speech from silent lip movement, which is called lip reading, is a challenging task due to 1) the inherent information insufficiency of lip movement to fully represent the speech, and 2) the existence of homophenes that have similar lip movement with different pronunciations. In this paper, we try to alleviate the aforementioned two challenges in lip reading by proposing a Multi-head Visual-audio Memory (MVM). Firstly, MVM is trained with audio-visual datasets and remembers audio representations by modelling the inter-relationships of paired audio-visual representations. At the inference stage, visual input alone can extract the saved audio representation from the memory by examining the learned inter-relationships. Therefore, the lip reading model can complement the insufficient visual information with the extracted audio representations. Secondly, MVM is composed of multi-head key memories for saving visual features and one value memory for saving audio knowledge, which is designed to distinguish the homophenes. With the multi-head key memories, MVM extracts possible candidate audio features from the memory, which allows the lip reading model to consider the possibility of which pronunciations can be represented from the input lip movement. This also can be viewed as an explicit implementation of the one-to-many mapping of viseme-to-phoneme. Moreover, MVM is employed in multi-temporal levels to consider the context when retrieving the memory and distinguish the homophenes. Extensive experimental results verify the effectiveness of the proposed method in lip reading and in distinguishing the homophenes.

\end{abstract}

\section{Introduction}
Lip reading, also known as Visual Speech Recognition (VSR), is a task that recognizes speech by watching lip movements only. It has a wide range of positive applications such as conversation with people who cannot make a voice (\eg, aphonia), an auxiliary technology for audio-based speech recognition in a noisy environment, and video conference in a crowded or silent environment. With the diverse applications, it has drawn big attention for a long time. However, compared to audio-based Automatic Speech Recognition (ASR), it is still regarded as a challenging task due to the following two reasons. First is the inherent information insufficiency of lip movements. The speech is jointly produced with various human organs such as vocal folds, larynx, tongue, and lips \cite{sataloff1992human}. Therefore, watching lip movements individually might be insufficient to fully represent the speech. The second is the existence of homophenes. The homophenes refer to some different pronunciations that show the same lip movements. Therefore, the same lip movements can be observed from different pronunciations. In other words, the viseme-to-phoneme mapping is one-to-many which places a challenge on lip reading. 

Recently, deep memory network \cite{weston2014memory,sukhbaatar2015end} that remembers valuable information learned during training and exploits the stored knowledge on a given task shows powerful performances on diverse applications such as question answering \cite{miller2016keyvalue}, video prediction \cite{lee2021videopredictionmem}, and retrieval \cite{song2018memoryforcrossmodalretrieval,huang2019acmmforfew-shotmatching}. Especially, a cross-modal memory network \cite{zhang2020fewshotcrossmodalmem} that saves different modal information in distinct memories shows promising results on information-limited situation such as few-shot learning scenario. By using a cross-modal memory network, it is possible to read one modal feature from another modal feature as being inputs, which is a highly attractive characteristic when applied to a task where only a single modal input is available. Therefore, the cross-modal memory seems valuable to be extended to the lip reading task using the audio-visual modalities \cite{kim2021visualaudiomem}, and to fulfill insufficient information of lip movement with the saved audio information read from the memory. However, directly applying the cross-modal memory into lip reading is not trivial due to the existence of homophenes. Since the visual features of homophenes are similar, they might point to the same value memory slots when addressing the memory. In other words, the same audio features could be obtained by different lip movement videos of homophenes, preventing the lip reading system from fully utilizing the audio information. Therefore, in order to take full advantage of the cross-modal memory, a method of considering the homophenes while bringing the audio information through memory is necessary.

In this paper, we propose Multi-head Visual-audio Memory (MVM) for lip reading, to mitigate the two aforementioned challenges, homophenes and information insufficiency of lip movements. The proposed visual-audio memory consists of multi-head key memories and one value memory. The value memory saves representative audio features and the key memory saves visual features utilized to retrieve proper audio representations from the value memory. Similar to the multi-head attention proposed in \cite{vaswani2017attention} that attends to information at different positions, MVM allows the model to jointly consider the information from different possible audio representations, even if the visual features of homophenes are given. Moreover, we employ the proposed MVM in multi-temporal levels so that the context can be considered, when querying the key memories to find out proper audio features from homophenes. With the proposed MVM, the lip reading network can fully utilize audio information stored in the value memory with an enhanced ability to distinguish the homophenes. 

The effectiveness of the proposed MVM is validated on popular benchmark databases. Moreover, we analyze and demonstrate the effectiveness of the proposed MVM in distinguishing homophenes by examining the pairs of words that belong to homophenes and visualizing the addressing scores of different head key memories. 

The contributions of this work are summarized as follows:
\begin{itemize}
    \item In order to consider the homophenes while bringing the saved audio information through a memory network, we propose Multi-head Visual-audio Memory (MVM) network that can extract possible candidate audio representations from one visual lip movement.
    \item To capture the context while querying the key memory and retrieving the saved audio features in the value memory, we employ the memory in multi-temporal levels so that the ambiguous mapping of viseme-to-phoneme can be refined.
    \item The proposed MVM achieves state-of-the-art performances on word-level lip reading in both English and Mandarin. Moreover, we analyze and validate the effectiveness of MVM in distinguishing homophenes.
\end{itemize}

%------------------------------------ Figure 1
%#################################################
\begin{figure*}[t!]
	\begin{minipage}[b]{1.0\linewidth}
		\centering
		\centerline{\includegraphics[width=18cm]{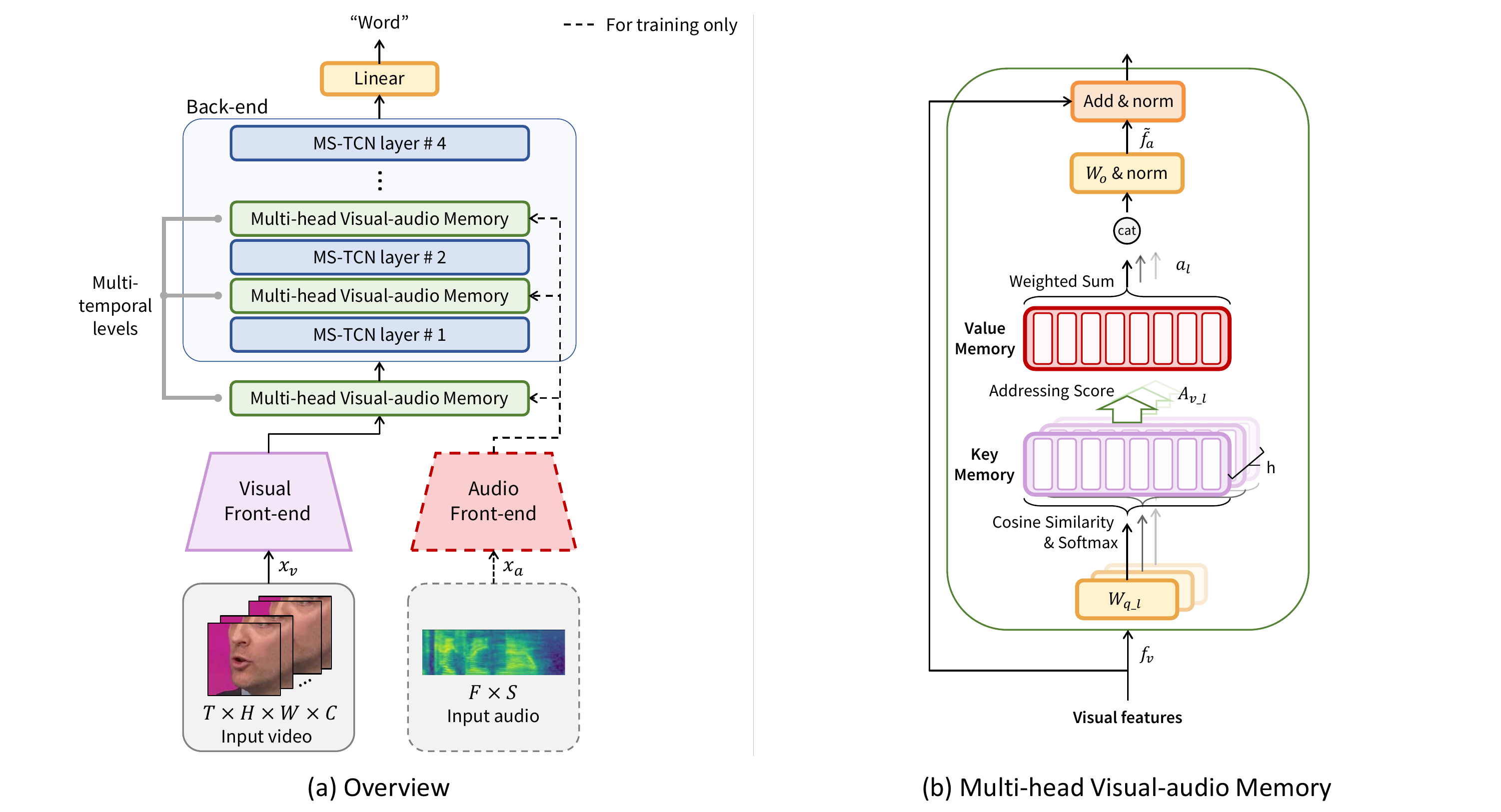}}
	\end{minipage}
	\caption{Illustration of the proposed method. (a) Overview of the proposed lip reading framework on word-level lip reading. During training, audio information is saved into the proposed memory and the learned knowledge is utilized at the inference stage without using the audio input. (b) Multi-head Visual-audio Memory (MVM). With the $h$ head key memories and a single value memory, it can extract different possible saved audio representations from the value memory with one visual feature.}
	\label{fig:1}
\end{figure*}
%##################################################

\section{Related Work}
\subsection{Lip Reading}
Lip reading \cite{chung2016lrw,ma2021denselyconnected,akbari2018lip2audspec,kim2021vcagan} is a task that recognizes speech from lip movements. Many research contributions in lip reading have focused on finding better spatio-temporal features for modelling visemes, the counterpart of phonemes in speech audio, from lip video. \cite{petridis2017resnetlstm} successfully recognized words by using front-end architecture of one 3D convolution layer and 2D ResNet \cite{he2016resnet} and back-end composed of LSTM. To capture the lip movement better, some studies \cite{weng2019twostream, xiao2020deformation} proposed two-stream networks that jointly model the raw video and the optical flow. \cite{martinez2020mstcn} improved the temporal encoding by proposing Multi-Scale Temporal Convolutional Network (MS-TCN) and boosted the word-level lip reading performance. Along with the advances in word-level lip reading, sentence-level VSR has also made great progress. \cite{assael2016lipnet} proposed the first deep learning based end-to-end sentence-level lip reading model using Connectionist Temporal Classification (CTC) \cite{graves2006ctc} loss function. \cite {chung2017lrs2} extended it to unconstrained sentence-level lip reading, proposing an English audio-visual corpus dataset and a Seq2Seq-based architecture \cite{cho2014seq2seq}. \cite{afouras2018deep} further improved the sentence-level lip reading performance by using Transformer \cite{vaswani2017attention} that has shown superior performances in language processing.

Apart from the architectural improvement, some studies have focused on utilizing audio information to complement the visual information for lip reading. \cite{afouras2020asrisall} tackled a problem that the labeled audio-visual dataset is much smaller than that of unlabeled. They proposed a method that utilizes the unlabeled data using a pre-trained ASR model. They trained a lip reading model with large-scale unlabeled data by guiding to follow the prediction of a pre-trained ASR model. On a similar line, \cite{zhao2020hearing, ren2021learningfromthemaster} proposed to train the lip reading network using teacher-student framework. They utilized a pre-trained ASR model as a teacher and trained the lip reading model as a student using knowledge distillation \cite{hinton2015distilling}. By distilling the knowledge of the teacher model into the student model, the lip reading model is expected to learn better representations for speech recognition. \cite{ma2021lira} proposed a self-supervised learning method to learn powerful visual speech representations. They pre-trained the visual front-end to predict corresponding acoustic features from the lip video, then fine-tuned with lip reading task loss. They showed its effectiveness in learning discriminative visual representations by achieving impressive lip reading performances. \cite{kim2021visualaudiomem,hong2021speechmem} proposed to utilize a cross-modal memory network to save and extract the audio representations, and \cite{kim2021cromm} described a limitation of cross-modal memory on handling the homophenes in lip reading.

In this paper, we also try to complement the visual information of lip movement using audio information. We employ a cross-modal memory network that memorizes visual and audio representations in the different memories (\ie, key and value), namely visual-audio memory. 
Apart from the previous methods, to handle the homophenes when reading the lips, we extend the cross-modal memory network to consist of multi-head key memories, and consider the context by applying it in multi-temporal levels. With the enriched context and candidate audio representations, the proposed method can effectively distinguish the homophene words.

\subsection{Memory Network}
Memory network is originally designed to alleviate the problem of forgetting information mainly in sequential data modelling. By augmenting the neural network with an external memory \cite{weston2014memory,sukhbaatar2015endmem}, it can flexibly utilize the important information learned during training. Due to its effectiveness, it has been widely applied not only in sequence modelling but also in question-answering \cite{miller2016keyvalue}, visual explanation \cite{kim2021mcam}, few-shot learning \cite{zhu2018fewshotmem}, pedestrian detection \cite{kim2021pedemem}, and video prediction \cite{lee2021videopredictionmem}. The memory network is also used for cross-modal data. \cite{song2018memoryforcrossmodalretrieval} proposed a memory network that saves multi-modal representations for cross-modal retrieval. It pre-learns and stores the features with semantic concepts of each modal. When a query is given, it searches supporting clues in different modalities and aggregates the clues for retrieval. \cite{zhang2020fewshotcrossmodalmem} applied multi-modal memory network for few-shot activity recognition. Their cross-modal memory is read and written using other modal features, which guarantees each memory slot to correspond with each other.

Distinct from the previous methods, we propose Multi-head Visual-audio Memory (MVM) based on cross-modal memory that saves visual and audio modalities to alleviate the ambiguity mapping of viseme-to-phoneme in lip reading. In the proposed MVM, multi-head key memories enable the lip reading network to consider the possible candidate audio representations for a given visual lip movement.

\section{Proposed Method}
Let $x_v \in \mathbb{R}^{T \times H \times W \times C}$ be a lip video with frames of $T$, height of $H$, width of $W$, and color channel size of $C$, $x_a \in \mathbb{R}^{F\times S}$ be a mel-spectrogram converted from speech audio corresponded to the lip video, where $F$ represents mel spectral size and $S$ is frame length, and $y$ be the corresponding ground-truth labels. The input video $x_v$ and audio $x_a$ are embedded through each respective front-end module as follows: $f_v=E_v(x_v)\in\mathbb{R}^{T\times D}$ and $f_a=E_a(x_a)\in\mathbb{R}^{T\times D}$, where $f_v$ and $f_a$ represent embedded visual and audio features, respectively, $E_v(\cdot)$ and $E_a(\cdot)$ are the visual and audio front-end modules, and $D$ is the dimension of embedding. Note since the paired audio and video are supposed to be aligned in time with the same duration, the front-end modules can be designed to output having the same frame numbers (\ie, $T$). Then, our objective is to save the visual features $f_v$ and the audio features $f_a$ in distinct memory networks, while learning the inter-relationships of the two modalities in training procedure. At the inference stage where input lip video is available only, we can extract the saved audio features from the memory by examining the learned inter-relationships using the input visual features. 
The overview of the proposed lip reading framework is illustrated in Fig. \ref{fig:1}a. Since there exist homophenes that have the same lip movement with different pronunciations, the mapping of viseme-to-phoneme is not one-to-one mapping but one-to-many. Thus, we explicitly design the cross-modal memory to represent the inter-relationships of one-to-many mapping to correctly extract the saved audio features by proposing MVM.

\subsection{Visual-audio Memory Network}
Visual-audio memory network, based on a cross-modal memory, stores visual features in key memory $M_v \in \mathbb{R}^{N\times D}$ and the audio features in value memory $M_a \in \mathbb{R}^{N\times D}$, where $N$ is the number of memory slots. Since the key memory and value memory are trained to save and to read the features of paired audio-visual data, it is possible to obtain the saved audio features by using visual inputs only. Our visual-audio memory operates with similarity-based reading (\ie, addressing) and writing (\ie, saving) \cite{kim2021visualaudiomem, lee2021videopredictionmem}. When a visual feature $f_v$ is given, the addressing score of $i$-th memory slot for $j$-th frame is obtained as follows,
\begin{align}
\label{eq:1}
    \mathcal{A}_v^{i,j} = \frac{\exp(\alpha\cdot d(M_v^i,W_q^\top f_v^j))}{\sum_{m=1}^N \exp(\alpha\cdot d(M_v^m,W_q^\top f_v^j))},
\end{align}
where $d(\cdot)$ is a cosine similarity metric, $W_q \in \mathbb{R}^{D \times D}$ represents projection weight for querying, and $\alpha$ is a scaling factor. The addressing score represents which memory slot contains the most relevant features to the given query visual feature. With the addressing score $\mathcal{A}_v^{i,j}$ obtained by using key memory and visual features, the saved audio features are extracted from the value memory as follows,
\begin{align}
    a^j = \sum_{i=1}^N\mathcal{A}_v^{i,j} \cdot M_a^i,
\end{align}
\begin{align}
    \tilde{f}_a^j = W_o^\top a^j,
\end{align}
where $a^j$ is the extracted audio features from the value memory and $W_o\in \mathbb{R}^{D\times D}$ represents projection weights. Therefore, besides the input visual features $f_v$, the lip reading model can additionally utilize the audio knowledge features $\Tilde{f}_a=\{\Tilde{f}_a^j\}_{j=1}^T$. By jointly modelling the features of two modalities, the insufficient information of lip movements can be complemented with the audio information.

\subsection{Multi-head Visual-audio Memory Network}
\label{sec:3.2}
Even though the visual-audio memory can bring the audio information without using the audio inputs during inference, it could fail on accessing proper memory slots if lip videos of homophenes are presented. This is natural since the visual-audio memory finds the saved audio using a fixed key memory and a given visual feature, and the visual features of homophene lip videos are similar to each other. In order to mitigate the ambiguity of the viseme-to-phoneme mapping, we strengthen the visual-audio memory to Multi-head Visual-audio Memory (MVM) with a multi-head structure. Specifically, the number of key memories (\ie, heads) is increased to $h$ (\ie, number of heads) while that of the value memory stays one \cite{lample2019mhmattention} as shown in Fig. \ref{fig:1}b. With the $h$ heads, the memory network can extract possible candidate audio representations. Similar to the multi-head attention \cite{vaswani2017attention}, it can be interpreted that the $h$ different audio representations obtained from the value memory allow the lip reading network to jointly consider the possible audio information for a given lip movement. Moreover, this can also be viewed as a realization of one-to-many mapping that of viseme-to-phoneme mapping, since MVM outputs $h$ saved audio features with a given visual feature.

The multi-head key memory is defined with $M_v=\{M_{v\_1}, ..., M_{v\_h}\}$, where $M_{v\_l}\in\mathbb{R}^{N\times \frac{D}{h}}$ represents $l$-th head key memory. Then, the addressing score of $i$-th memory slot for $j$-th frame is obtained for each head key memory as follows,
\begin{align}
    \mathcal{A}_{v\_l}^{i,j} = \frac{\exp(\alpha\cdot d(M_{v\_l}^i,W_{q\_l}^\top f_v^j))}{\sum_{m=1}^N \exp(\alpha\cdot d(M_{v\_l}^m,W_{q\_l}^\top f_v^j))},
\end{align}
where $W_{q\_l}\in\mathbb{R}^{D\times\frac{D}{h}}$ represents projection weight for $l$-th head. With the addressing score, the $h$ different audio features are extracted from the value memory and aggregated.
\begin{align}
    a_l^j = \sum_{i=1}^N\mathcal{A}_{v\_l}^{i,j} \cdot M_a^i,
\end{align}
\begin{align}
    \Tilde{f}_{a}^j = W_o^\top\text{Concat}(a_1^j,...,a_h^j),
\end{align}
where $a_l^j$ represents extracted audio features from the value memory using $l$-th head key memory and $W_o\in\mathbb{R}^{Dh \times D}$ is embedding weight that aggregates the $h$ different extracted audio features. The audio knowledge features $\tilde{f}_a$ are fused with the visual features $f_v$ by addition, followed by a layer normalization \cite{ba2016layernorm} that we omit in the equations for simple notations. Then, the back-end module models the context for prediction using the fused representations of visual and audio knowledge features.

%------------------------------------ Figure 2
%#################################################
\begin{figure}[t!]
	\begin{minipage}[b]{1.0\linewidth}
		\centering
		\centerline{\includegraphics[width=5.6cm]{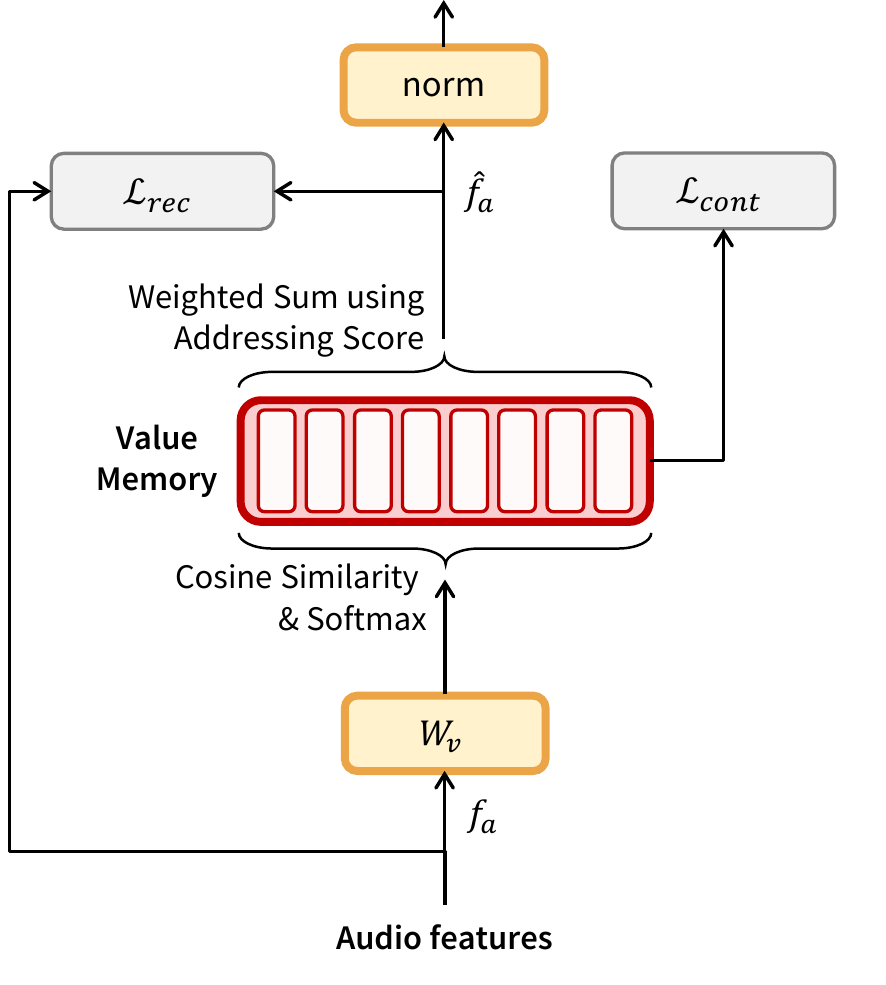}}
	\end{minipage}
	\caption{Learning to save audio representations into value memory with reconstruction and contrastive losses.}
	\label{fig:2}
\end{figure}
%##################################################

\subsection{Training and Inference}
MVM is trained along with the whole lip reading network in an end-to-end manner. In order to save the representative audio representations in the value memory, we employ reconstruction-based learning and contrastive learning to train the value memory. The reconstruction-based learning guarantees the saved representations in the value memory correctly contain the audio information. It is defined with a cosine similarity-based reconstruction loss as follows,
\begin{align}
    \mathcal{L}_{rec}= ||1 - d(\hat{f}_a,f_a)||_1,
\end{align}
where $\hat{f}^j_a=\sum^N_{i=1}\mathcal{A}_a^{i,j}\cdot M_a^i$ represents reconstructed audio features from the value memory $M_a$ by using the addressing score $\mathcal{A}_a^{i,j}$. Note that $\mathcal{A}_a^{i,j}$ is obtained similar to Eq. \ref{eq:1} by substituting key memory $M_v$ and visual feature $f_v$ with value memory $M_a$ and audio feature $f_a$. In addition, to save the representative audio features that are distinct as possible from each other, we propose to use contrastive learning on the value memory as follows,
\begin{align}
    \mathcal{L}_{cont}= \sum_{i\neq j} ||d(M_a^i,M_a^j)||_1.
\end{align}
The contrastive loss, $\mathcal{L}_{cont}$, guides the different memory slots to have less similar audio features, which leads the value memory can contain discriminative audio representations. Fig. \ref{fig:2} shows the training process of the value memory.

During training, the key memory $M_v$ is trained to attend and extract proper audio features saved in the value memory with lip reading loss such as cross-entropy loss and CTC loss \cite{graves2006ctc}. The task loss is applied as follows,
\begin{align}
    \mathcal{L}_{task} = \mathcal{F}(g(f_v + \tilde{f}_a), y) + \mathcal{F}(g(f_v + \hat{f}_a), y),
\end{align}
where $\mathcal{F}$ represents the lip reading loss and $g$ represents the back-end module. The first term of the task loss guarantees the classification performance using both visual features $f_v$ and audio knowledge features $\tilde{f}_a$ obtained from the value memory, which is straightly the inference form of the proposed method. The second term makes possible the end-to-end training by guiding the audio front-end module to extract useful audio features (which will be saved into the value memory) for speech recognition.
The total loss function is the sum of the pre-defined loss functions,
\begin{align}
    \mathcal{L}_{tot} = \mathcal{L}_{task} + \mathcal{L}_{rec} + \mathcal{L}_{cont}.
\end{align}

During inference, as the value memory is trained to save the representative audio features and the key memory is trained to extract proper saved audio features from the value memory, we do not require the input audio. Thus, the audio front-end is detached and only the input video is utilized.

\subsection{Considering Context in Multi-temporal Levels}
By implementing the one-to-many mapping through the proposed MVM, we can obtain the possible audio representations corresponding to an input lip visual representation. However, there is still room for improvement in terms of the clarity of the mapping. Since the visual features $f_v$ are embedded using a visual front-end module that usually has fixed temporal receptive field, $f_v$ also has limited temporal information. For example, the popular front-end module \cite{petridis2017resnetlstm} in lip reading has a temporal receptive field of 5 consecutive frames. Therefore, when querying the key memory by using the visual features, it might fail to consider the context beyond 5 frames which is another key for distinguishing the homophenes. In order to improve the memory addressing stage, we employ MVM in multi-temporal levels, so that the context at different temporal ranges can be considered when querying the key memory. To this end, MVM is applied not only between the front-end and back-end but also in the back-end that models temporal context, in a multi-layer fashion, as shown in Fig. \ref{fig:1}a. Thus, the mapping of viseme-to-phoneme can be elaborated as the layers go deeper by considering the context.

\section{Experiments}

\begin{table}[]
	\renewcommand{\arraystretch}{1.2}
\centering
\resizebox{0.99\linewidth}{!}{
\begin{tabular}{ccccc}
\hline \hline
\multicolumn{4}{c}{\textbf{Method}} & \\ \cmidrule{1-4}
\textbf{Baseline} & \textbf{\makecell{Visual-audio\\Memory}} & \textbf{\makecell{Multi-head\\Visual-audio\\Memory}} & \textbf{\makecell{Multi-temporal\\Level}}  & \textbf{ACC(\%)} \\ \hline
\cmark & \xmark & \xmark & \xmark & 86.1 \\
\cmark & \cmark & \xmark & \xmark & 86.9 \\
\cmark & \cmark & \cmark & \xmark & 87.2 \\ \hdashline
\cmark & \cmark & \cmark & \cmark & \textbf{88.5} \\ \hline \hline
\end{tabular}}
\caption{Contributions of each proposed component}
\label{table:1}
\end{table}

\subsection{Dataset}
We conduct the experiments on both word- and sentence-level lip reading databases. For the word-level lip reading, we use LRW \cite{chung2016lrw} and LRW-1000 \cite{yang2019lrw1000} datasets, which are in English and Mandarin, respectively. For the sentence-level lip reading, we use LRS2 \cite{chung2017lrs2} dataset.

The LRW is an English word-level lip reading dataset which includes 500 words with a maximum of 1,000 training videos each. We crop the video into 136$\times$136 centered at the lip without aligning the face and resize it into 112$\times$112. The audio is transformed into a mel-spectrogram using a window size of 400, hop size of 160, and 80 mel-filters.

The LRW-1000 is a word-level dataset in Mandarin. It contains a total of 718,018 videos with 1,000 words. The dataset provided is already cropped, thus we resize the video into 112$\times$112 without cropping. Since the audio provided is longer than the video, we use 0.2 sec longer video back and forth to match the duration of video and audio.

The LRS2 is a sentence-level audio-visual dataset. It is collected from British news programs to form 224 hours videos, containing large variations in head pose and illumination. The video is preprocessed similar to LRW.

\subsection{Implementation Details}
For the visual front-end, we use a popular architecture in lip reading that consists of one 3D convolution layer and ResNet-18 \cite{petridis2017resnetlstm}. The audio front-end is designed with 2 convolution layers with stride 2 and one Residual block. For the back-end, we use MS-TCN \cite{martinez2020mstcn}, a state-of-the-art architecture, for word-level lip reading and Transformer \cite{vaswani2017attention} architecture for sentence-level lip reading following \cite{afouras2018deep}. MVM is empirically designed with 8 heads ($h$) and 112 slots ($N$). For the word-level lip reading, it is applied in 4 different levels, before the back-end and after every MS-TCN layer except the last layer. For the sentence-level lip reading, MVM is applied in 4 different levels, after the front-end, and before the 2nd, 4th, and 6th transformer encoder layer.

For training the word-level lip reading, data augmentation including random horizontal flipping, random erasing, and mixup \cite{zhang2017mixup, ma2021bornagain} is applied. The cross-entropy loss is utilized for the lip reading loss function. For sentence-level lip reading training, we follow the training schemes of \cite{zhang2019spatio, chung2017lrs2, afouras2018deep} that 1) pre-train the visual front-end on the LRW dataset, 2) pre-train the whole network using pre-train sets of the LRS2 and the LRS3 with curriculum setting, and 3) finally train and test on the LRS2 dataset. The hybrid CTC/Attention \cite{watanabe2017hybrid} loss function is utilized. The external language model is not utilized for decoding stage. We use AdamW optimizer \cite{loshchilov2017adamw}, batch size of 200, 64, and 40 for LRW, LRW-1000, and LRS2, respectively, with initial learning rate of 0.0001, and $\alpha$ is set to 16. We use four Titan RTX GPUs (24GB) and Intel Xeon Gold 6130 CPU.

\subsection{Experimental Results}
\subsubsection{Ablation study}
In order to examine the contributions of each proposed component on lip reading, we build 4 variants of the model, 1) \textit{baseline} lip reading model without cross-modal memory, 2) lip reading model \textit{with cross-modal memory} based on \cite{kim2021visualaudiomem}, 3) lip reading model with MVM and \textit{without applying in multi-temporal levels}, and 4) the final proposed model with MVM \textit{in multi-temporal levels}. Table \ref{table:1} shows the word-level accuracy of each model on the LRW dataset. The baseline model which does not contain the visual-audio memory achieves 86.1\% word accuracy. By applying the visual-audio memory to the baseline, the accuracy is improved to 86.9\%, and this result confirms that bringing the audio information can complement the insufficient information of visual lip movement in speech recognition. By considering the homophenes with extracting possible audio features using the proposed MVM, we can improve the performance to 87.2\%. It is valuable to note that the key and value memory parameters remain similar before and after the multi-head key is applied, as we reduce the dimension of key memory according to the head size. Finally, by applying MVM in multi-temporal levels, the word-accuracy achieves 88.5\% with additional 1.3\% improvement. The result clearly shows that applying MVM in multi-temporal levels is beneficial by considering the context when retrieving the saved audio feature.

\begin{table}[]
	\renewcommand{\arraystretch}{1.2}
	\renewcommand{\tabcolsep}{7mm}
\centering
\resizebox{0.99\linewidth}{!}{
\begin{tabular}{c|c|c|c}
\hline \hline
\multicolumn{4}{c}{\textbf{\# Memory slot ($N$)}} \\ \hline
\textbf{28} & \textbf{56} & \textbf{112}* & \textbf{224} \\ \hline
88.02\% & 88.21\% & 88.51\% & 88.06\% \\ \hline \hline
\multicolumn{4}{c}{\textbf{\# Head ($h$)}} \\ \hline
\textbf{1} & \textbf{4} & \textbf{8}* & \textbf{16} \\ \hline
87.53\% & 88.18\% & 88.51\% & 88.30\% \\ \hline \hline
\multicolumn{4}{c}{\textbf{\# Multi-temporal level}} \\ \hline
\textbf{1} & \textbf{2} & \textbf{3} & \textbf{4}* \\ \hline
87.22\% & 87.44\% & 88.00\% & 88.51\% \\ \hline \hline
\end{tabular}}
\caption{Effect of different hyperparameters of MVM.}
\label{table:2}
\end{table}

\begin{table}[t!]
\renewcommand{\arraystretch}{1.4}
\resizebox{0.9999\linewidth}{!}{
\begin{tabular}{c|c|c}

\hline \hline
\textbf{Method} & \textbf{Backbone / Method} & \textbf{ACC (\%)} \\ \hline
\citet{luo2020pcpg} & R18 + BiGRU + GRU & 83.5 \\
\citet{weng2019twostream} & T I3D + BiLSTM & 84.1 \\
\citet{xiao2020deformation} & T R18 + BiGRU & 84.1 \\
\citet{zhao2020mimaximization} & R18 + BiGRU + LSTM & 84.4 \\
\citet{xu2020discriminative} & P3D R50 + BiLSTM & 84.8 \\
\citet{zhang2020facecutout} & R18 + BiGRU / Face Cutout & 85.0 \\
\citet{martinez2020mstcn}   & R18 + MS-TCN & 85.3 \\
\citet{kim2021visualaudiomem} & R18 + BiGRU / Mem & 85.4 \\
\citet{ma2021bornagain}   & R18 + MS-TCN / Born-Again & 87.9 \\ 
\citet{ma2021lira} & R18 + MS-TCN / LiRA & 88.1 \\
\citet{ma2021denselyconnected} & R18 + DC-TCN & 88.4 \\ \hline
\textbf{Proposed Method} & R18 + MS-TCN / MVM & \textbf{88.5} \\ \hline \hline
\end{tabular}}
\caption{Word accuracy comparison on LRW. 
}
\label{table:3}
\end{table}

In order to examine the performance changes according to the hyperparameter of MVM, we experiment by differing the memory slots ($N$), the number of heads ($h$), and the number of multi-temporal levels on the LRW dataset. Table \ref{table:2} shows the experimental results obtained under different hyperparameters. To examine the effect of each hyperparameter, we fix the other factors when changing one factor and the fixed number is represented in the table with a star mark (*). The experimental result shows that too small number of memory slots are less effective, which means the audio representations cannot be fully covered with a small number of memory slots. Moreover, assigning an excessive number of memory slots does not lead to an increase in performance. By differing the number of head key memories, we observe that more heads tend to be beneficial to lip reading by providing possible audio information for the homophenes. Finally, we can further improve the mapping of viseme-to-phoneme by applying MVM at multi-temporal levels in the back-end module. We use memory slots ($N$) of 112, the number of heads ($h$) of 8, and 4 multi-temporal levels in the other experiments.

\subsubsection{Comparison with state-of-the-art methods}

In order to verify the effectiveness of the proposed method, we compare the word-level lip reading performances with the state-of-the-art methods. Table \ref{table:3} shows the comparison of word-level lip reading performances on the LRW dataset. The proposed method achieves 88.5\% word accuracy and sets a new state-of-the-art performance. The result shows the effectiveness of the proposed method on modelling the visual representations and complementing the insufficient information with the audio information. In addition, the word-level lip reading comparison on the LRW-1000 dataset is shown in Table \ref{table:4}. The LRW-1000 dataset is a relatively challenging dataset than the LRW due to its unbalanced training samples. Even in the challenging environment, the proposed method outperforms the previous state-of-the-art method \cite{kim2021visualaudiomem} by 3\% accuracy, achieving 53.82\% word accuracy. The two word-level lip reading results confirm that the proposed method is beneficial for lip reading by 1) fulfilling the insufficient visual information with the saved audio information during training and 2) considering the homophenes using multi-head key memory and context in multi-levels.

\begin{table}[t!]
\renewcommand{\arraystretch}{1.4}
\resizebox{0.9999\linewidth}{!}{
\begin{tabular}{c|c|c}
\hline \hline
\textbf{Method} & \textbf{Backbone / Method} & \textbf{ACC (\%)} \\ \hline
\citet{yang2019lrw1000} & R34 + BiGRU & 38.19 \\
\citet{luo2020pcpg} & R18 + BiGRU + GRU & 38.70 \\
\citet{zhao2020mimaximization} & R18 + BiGRU + LSTM & 38.79 \\
\citet{martinez2020mstcn}   & R18 + MS-TCN & 41.40 \\
\citet{xiao2020deformation} & T R18 + BiGRU & 41.93 \\
\citet{ma2021denselyconnected} & R18 + DC-TCN & 43.65 \\ 
\citet{zhang2020facecutout} & R18 + BiGRU / Face Cutout & 45.24 \\
\citet{ma2021bornagain}   & R18 + MS-TCN / Born-Again & 46.60 \\  
\citet{kim2021visualaudiomem} & R18 + BiGRU / Mem & 50.82 \\ \hline
\textbf{Proposed Method} & R18 + MS-TCN / MVM & \textbf{53.82} \\ \hline\hline 
\end{tabular}}
\caption{Word accuracy comparison on LRW-1000. 
}
\label{table:4}
\end{table}

\begin{table}[t!]
	\renewcommand{\arraystretch}{1.4}
	\renewcommand{\tabcolsep}{10mm}
\centering
\resizebox{0.95\linewidth}{!}{
\begin{tabular}{cc}
\hline \hline
\textbf{Method} & \textbf{WER(\%)} \\ \hline
Baseline \cite{afouras2018deep} & 49.8 \\
\textbf{Proposed Method} & \textbf{44.5}         \\ \hline \hline
\end{tabular}}
\caption{WER comparison with a baseline model on LRS2.}
\label{table:5}
\end{table}

Moreover, we compare the proposed method on sentence-level lip reading using the LRS2 dataset with the baseline model that has the same architecture as the proposed method except for MVM. Table \ref{table:5} shows the Word Error Rate (WER) performance of each method. We can clearly observe the effectiveness of MVM as it improves the performance by 5.3\% WER from the baseline \cite{afouras2018deep}. Consistent with the two word-level lip reading results, we can confirm the benefits of the proposed method in lip reading by improving the performance with a large gap.

\subsubsection{Effectiveness on distinguishing homophenes}

\begin{table}[t!]
\renewcommand{\arraystretch}{1.4}
\renewcommand{\tabcolsep}{7mm}
\resizebox{0.9999\linewidth}{!}{
\begin{tabular}{cc|cc}
\hline \hline
\multicolumn{4}{c}{Word accuracy (\%) / Performance change (\%)} \\ \hline
Living & Giving & Better & Matter \\
82 / {\textbf{+6}} & 76 / {\textbf{+12}} & 82 / {\textbf{+2}} & 70 / +0 \\ \hline
Million & Billion & Heard & Heart \\
84 / {\textbf{+2}} & 88 / {\textbf{+4}} & 72 / {\textbf{+8}} & 90 / {\textbf{+4}} \\ \hline 
Words & World & Black & Plans \\
66 / +0 & 76 / {\textbf{+4}} & 88 / {\textbf{+4}} & 88 / {\textbf{+2}} \\ \hline 
\hline
\end{tabular}}
\caption{Homophene word accuracy and its change compared to the baseline that contains one visual-audio memory.}
\label{table:6}
\end{table}

In order to analyze the effectiveness of the proposed method in distinguishing homophenes, we compare the word accuracy of the proposed method with a visual-audio memory model based on \cite{kim2021visualaudiomem} as a baseline. To this end, we extract the words list of the LRW dataset and find the paired words of homophenes based on the viseme-to-phoneme mapping table of \cite{cappelletta2012phonemetoviseme}. Table \ref{table:6} shows six pairs of words belonging to the homophenes and their predicted accuracy. Moreover, we represent the performance improvement by adopting the proposed method. For example, word accuracy for \textit{Giving} of the proposed method achieves 76\% and that of the baseline is 64\%. Therefore, the proposed method boosts the performance for \textit{Giving} by 12\%, which is denoted bold in the table. From the table, we can analyze the following two facts: 1) homophenes cause difficulty in lip reading, showing relatively less accuracy compared to the mean model accuracy (\ie, 88.5\%) and 2) by employing the proposed MVM within the multi-temporal levels, we can enhance the performance with a large gap from the model that utilizes visual-audio memory only, which mainly contributes to the improvement of mean model accuracy (\ie, 1.6\% from the baseline).

\begin{figure}[t!]
	\begin{minipage}[b]{1.0\linewidth}
		\centering
		\centerline{\includegraphics[width=8.6cm]{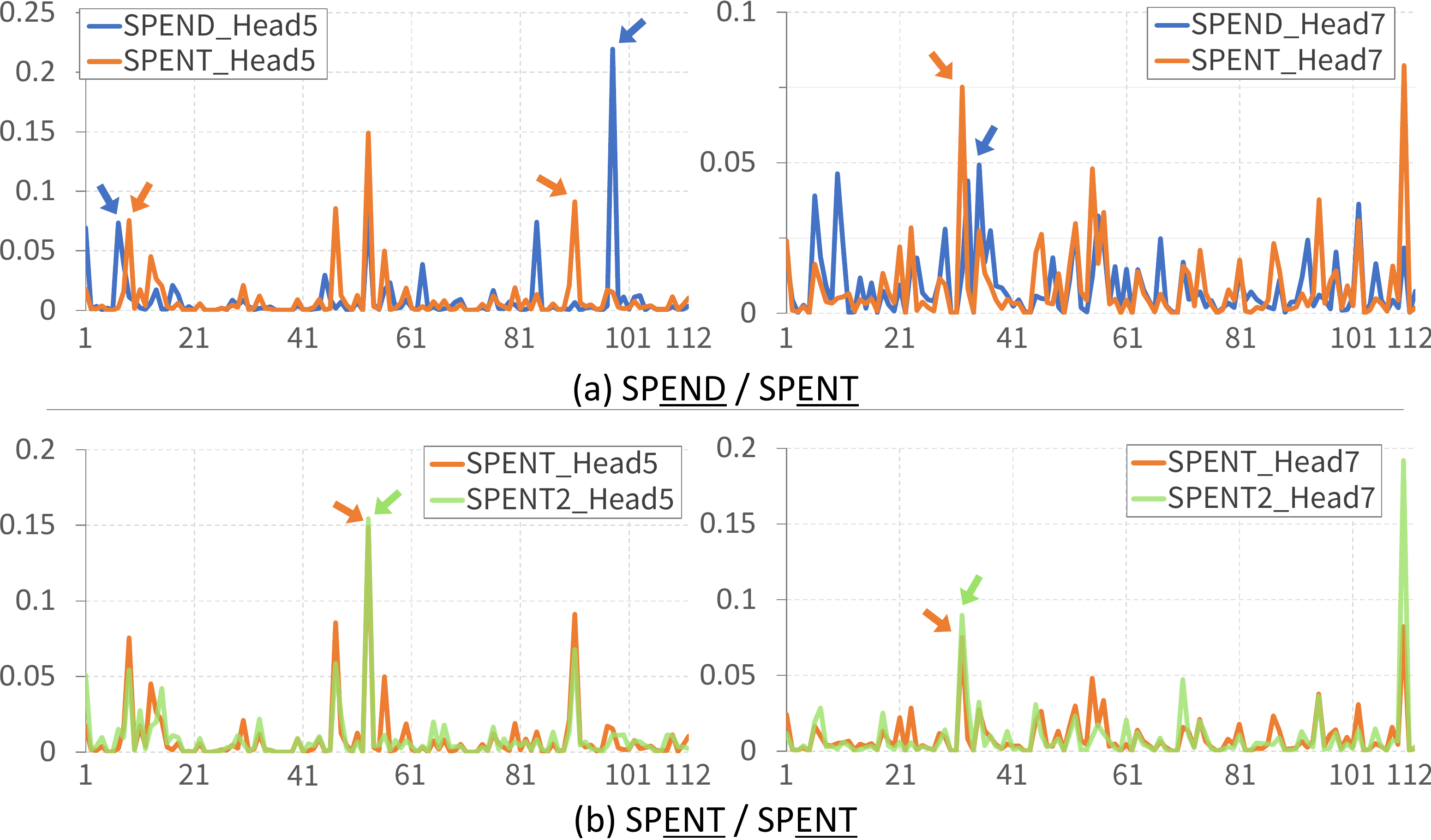}}
	\end{minipage}
	\caption{Addressing scores of head 5 and head 7 for (a) homophene words, and (b) the ones of the same words.}
	\label{fig:3}
\end{figure}

To verify whether MVM 1) extracts different audio representations and 2) distinguishes the homophene words, we visualize the addressing scores of different head key memories at the third level MVM for homophene words. Fig. \ref{fig:3}a shows addressing scores of head 5 and head 7 for homophene words \textit{END} and \textit{ENT} of \textit{SPEND} and \textit{SPENT}, respectively. It shows that the different value memory slots are accessed by different words, even if their lip movements are visually similar. This is because the context modeled at the back-end helps the memory retrieving stage to distinguish the homophenes. Moreover, we can find that the different heads attend different memory slots which means MVM provides possible candidate audio representations. Moreover, Fig. \ref{fig:3}b shows that the same word yield similar value memory addressing. From the visualization results, we can confirm the effectiveness of the proposed MVM on distinguishing the homophenes which induce challenges in lip reading.

%------------------------------------ Figure 3
%#################################################
%##################################################

\section{Conclusion}
In this paper, we have proposed Multi-head Visual-audio Memory (MVM) for lip reading. The proposed MVM can complement insufficient information of lip movement with audio knowledge when the audio input is not provided. Moreover, to distinguish the homophenes during memory accessing, MVM is designed with a one-to-many mapping fashion, which allows the lip reading network to jointly model the different possible audio representations that correspond to a given lip visual feature. Through extensive experiments, we have verified the effectiveness of the proposed framework in distinguishing homophenes.

\bibliography{aaai22}

\end{document}